\documentclass{article} 
\PassOptionsToPackage{numbers,sort&compress}{natbib}
\usepackage{iclr2026_conference,times}

\usepackage{amsmath,amsfonts,bm}

\def\eqref#1{equation~\ref{#1}}

\def\1{\bm{1}}

\DeclareMathAlphabet{\mathsfit}{\encodingdefault}{\sfdefault}{m}{sl}
\SetMathAlphabet{\mathsfit}{bold}{\encodingdefault}{\sfdefault}{bx}{n}

\usepackage{hyperref}
\usepackage{url}
\usepackage{graphicx}
\usepackage{wrapfig}
\usepackage{subcaption}
\usepackage{amsmath,amssymb}
\usepackage{algorithm}
\usepackage[noend]{algpseudocode}
\usepackage{booktabs}
\usepackage[table]{xcolor}
\usepackage{multirow}
\definecolor{pnrInk}{HTML}{17262E}
\definecolor{pnrDecode}{HTML}{0E7C7B}
\definecolor{pnrDeploy}{HTML}{D1495B}
\definecolor{pnrCronus}{HTML}{6A4C93}
\definecolor{pnrAccent}{HTML}{E8A22E}
\definecolor{pnrMute}{HTML}{7C8A91}
\definecolor{pnrGrid}{HTML}{D8DEE0}
\definecolor{pnrPaper}{HTML}{F5F2EC}
\definecolor{pnrDecodeT}{HTML}{CFE5E4}
\definecolor{pnrDeployT}{HTML}{F4D4D9}
\definecolor{pnrCronusT}{HTML}{E1D9EC}
\definecolor{pnrAccentT}{HTML}{F8E8C7}
\newcommand{\sig}[1]{\textcolor{pnrAccent}{\textbf{#1}}}      
\newcommand{\ns}[1]{\textcolor{pnrMute}{#1}}                   
\newcommand{\present}[1]{\textcolor{pnrDecode}{#1}}
\newcommand{\deployc}[1]{\textcolor{pnrDeploy}{#1}}

\graphicspath{{figs/}}
\setcitestyle{numbers,square,comma}
\usepackage{caption}
\title{Present but Not Remembered:\\ Auditing How Frozen VLAs Encode, Deploy, and Steer Visual History}

\iclrfinalcopy

\author{%
Chih-Ting Liao\thanks{Corresponding author and project lead. \texttt{mill.liao@unsw.edu.au}} \quad Xin Cao \\
University of New South Wales
}

\begin{document}
\maketitle
\lhead{Preprint.}

\begin{abstract}
A frozen vision-language-action model (VLA) is handed the recent past at every step, yet the
literature rushes to \emph{add} memory without first asking what the model already does with the
history it has. We supply that missing diagnosis on the \emph{time} axis: how a policy encodes
and deploys observations \emph{across} timesteps. This is precisely the axis held fixed by the
only prior cross-architecture mechanistic study of VLAs, which localizes the vision-language
\emph{fusion} axis \emph{within} a single frame (vision dominates, architecture-independently);
we localize the orthogonal temporal coordinate and reach the opposite structural verdict. Using
layer-resolved linear probing and interchange interventions, we find a three-layer dissociation.
(i)~Past-frame content is linearly \emph{decodable} at every depth. (ii)~Yet the information
\emph{unique} to history, beyond the current frame, is essentially zero, a ceiling established
over an $81$-configuration probe sweep in three models and both architecture families: the stored
history is a redundant copy of the present. (iii)~History is causally \emph{deployed} into the action only under near-total loss of
the current frame, and the readout severs its dependence on history by the middle of the
network, confirmed by a second, orthogonal intervention. Across architectures the encoding is
identical ($A_4\!\approx\!0$ in both) but the deployment regime \emph{flips}: under the
\emph{same} occlusion one model's reliance on history rises (fallback) while the other's falls
(standing use). We cast the four measurements as a reusable, training-free temporal-deployment
audit (Algorithm~\ref{alg:tda}), and its injection leg adds a third, \emph{causal}
architecture-conditional axis: in the fallback regime, re-supplying history is content-blind (it
neither repairs occlusion nor disambiguates), causally confirming the redundancy; in the standing
regime the \emph{same} injection steers the action toward the donor. Whether history is steerable
thus tracks the deployment regime, not whether the architecture encodes history: a single
fallback-vs-standing axis organizes all three of \emph{what} is encoded, \emph{when} it is
deployed, and \emph{whether} it can be steered. VLAs do not
forget the past; they were never built to treat it as separate from the present. The diagnosis
hands the memory-augmentation literature a concrete lesson: inject information that is
\emph{unique} to the past, not more of it.
\end{abstract}

\section{Introduction}
\label{sec:intro}

A growing body of work equips vision-language-action models (VLAs) with explicit memory
modules, retrieval buffers, recurrent latents, trajectory overlays, learned summaries, and
benchmarks whether the augmented policy \emph{can} use the past
\citep{memoryvla2025,memer2025,torne2025longcontext,contextvla2025,tracevla2024,cronusvla2025}. Every such
effort presupposes a diagnosis that, to our knowledge, no one has performed: \emph{on a frozen
multi-frame policy, is the past stored as an independent memory or merely as a redundant copy of
the present, and is it ever causally read into the action?} Without that baseline, \emph{add
memory} is a fix in search of a characterized fault. The answer is architecture-conditional and
counter-intuitive: under the \emph{same} occlusion one policy leans \emph{harder} on history while
another leans \emph{less}, and whether injected history can \emph{steer} the action flips with it,
though both encode the past identically.

We provide the missing diagnosis on the \textbf{time axis}. This is deliberately orthogonal to
the closest mechanistic work. \citet{grant2026notall}, the first cross-architecture
mechanistic study of VLAs, localizes which \emph{modality} drives the action, finding vision
dominates language, \emph{independently} of architecture, a question about space \emph{within
a single timestep}. We ask how \emph{temporal history across timesteps} is encoded and
deployed. The two are complementary halves of a single mechanistic picture; we show below that
where Grant's fusion axis is architecture-\emph{independent}, the time axis is
architecture-\emph{dependent}.

Our method has two legs on a frozen, open-loop policy: a linear \emph{decode} probe that asks
whether past-frame content is present in the representation, and an \emph{interchange}
intervention that transplants a donor's history activations and measures the resulting shift
in the action, a direct, layer-resolved test of causal deployment. Open-loop evaluation on a
small, frozen, hashed stimulus set is a deliberate choice: it is what enables interchange at
every block, a localization that closed-loop rollout cannot provide. The unit of evidence is a
controlled stimulus pair under causal intervention, not a behavioral episode, and we lean on
statistical rigor (multi-seed, permutation, bootstrap) rather than episode volume.

We report a three-layer dissociation, summarized by the paper's title.
\noindent\textbf{Present:} past-frame content is linearly decodable at every depth, peaking mid-network.
\noindent\textbf{Redundant:} the information \emph{unique} to history, the part of $t\!-\!1$ not
already in $t$, tops out at $\approx\!0.02$ over an $81$-configuration probe sweep and is
$\approx\!0$ across three models and both architecture families ($A_4/A_1\!\lesssim\!0.05$),
so the decodable history is almost entirely redundant with the present, which is the mechanistic
answer to why history is not deployed: there is almost nothing unique to deploy.
\noindent\textbf{Not remembered (fallback-deployed):} history is causally deployed only under
near-total current-frame loss (gap significant only for full occlusion; milder degradations not
significant), and the action readout no longer reads history past the middle of the network.
Across architectures the encoding is identical, yet both the deployment \emph{and} the
content-steerability sign-flip with the fallback-vs-standing regime, the paper's central surprise.

\noindent\textbf{Contributions.} (1)~The first mechanistic study of the \emph{time} axis of VLA
memory, with a history-unique decode metric ($A_4$) that separates retained past information from
current-frame redundancy, the temporal complement to the fusion-axis study of
\citet{grant2026notall}. (2)~A training-free, reusable \emph{temporal-deployment audit}
(Algorithm~\ref{alg:tda}) that profiles any frozen VLA: whether the past is present ($A_1$),
unique ($A_4$), deployed and where ($\Delta_\ell$, cutoff $\ell^\star$), its regime ($\sigma$),
and whether injection carries history \emph{content}. (3)~An \emph{architecture-conditional
steerability law}: encoding is identical across families ($A_4\!\approx\!0$ in both), yet both the
deployment direction \emph{and} content-steerability sign-flip with the fallback-vs-standing
regime, established causally by injection (inert in the fallback regime, steering in the standing
one). The design implication follows: useful memory must add information \emph{unique} to the
past, injected before the mid-network cutoff we localize.

\section{Related work}
\label{sec:related}

\noindent\textbf{Mechanistic interpretability of VLAs.}
The closest work localizes the \emph{within-frame} modality axis: \citet{grant2026notall} show,
across architectures, that vision dominates language in driving actions \emph{independently} of
architecture, with the time axis held fixed. Others steer within-frame motion features
\citep{haon2025steering,buurmeijer2026observing} or decode a \emph{forward} world model
\citep{molinari2025emergent}; none addresses the causal deployment of the \emph{past} frame. We
localize that orthogonal time axis and reach the opposite structural verdict, deployment is
architecture-\emph{dependent}, so the two lines jointly say the VLA mechanism is shared on the
modality axis but divergent on the time axis (Table~\ref{tab:positioning}).

\begin{table}[t]
\centering
\caption{\textbf{Positioning: VLA $\times$ mechanistic interpretability.} Prior work localizes the
\emph{within-frame} modality axis, steers within-frame motion features, or adds memory without a
diagnosis; we localize the orthogonal \emph{time} axis and characterize how history is encoded,
deployed, and steered, reaching an architecture-\emph{conditional} verdict.}
\label{tab:positioning}
\footnotesize
\setlength{\tabcolsep}{3pt}
\rowcolors{2}{pnrPaper}{white}
\begin{tabular}{@{}l p{1.9cm} p{2.4cm} p{1.6cm} p{3.3cm}@{}}
\toprule
\textbf{Work} & \textbf{Axis} & \textbf{Method} & \textbf{Scope} & \textbf{Finding}\\
\midrule
Grant 2026 & modality / fusion (within-frame) & SAE, injection, probes & 6 models, closed-loop & vision dominates, arch.-\emph{independent}\\
H\"aon 2025 & motion features (within-frame) & activation steering & 2 models, real robot & zero-shot behavioral steering\\
Buurmeijer 2026 & feature obs./control (within-frame) & probe $+$ linear interv. & closed-loop & feature observability/control\\
Memory VLAs$^{\dagger}$ & \emph{add} memory (no diagnosis) & new trained modules & closed-loop & augmented policy \emph{can} use past\\
\textbf{Ours} & \textbf{time / history across frames} & decode, interchange, knockout, inject (\emph{frozen}) & 2 regimes, open-loop & present-but-redundant, fallback-deployed (L6); arch.-\emph{conditional}; standing-only steerable\\
\bottomrule
\end{tabular}

{\footnotesize $^{\dagger}$MemoryVLA, MemER, TraceVLA, ContextVLA, CronusVLA
\citep{memoryvla2025,memer2025,tracevla2024,contextvla2025,cronusvla2025}.}
\end{table}

\noindent\textbf{Benchmarks and memorization.}
Long-horizon manipulation benchmarks such as LIBERO \citep{libero2023} and SimplerEnv
\citep{simplerenv2024} evaluate whether policies \emph{can} solve history-dependent tasks
behaviorally; they do not localize whether or where native policies read the past.
\citet{he2026memorization} show, complementarily, that diffusion policies often \emph{memorize}
training actions, a different sense in which the learned mapping leans on stored content. We
provide the upstream mechanistic account: on a frozen multi-frame policy, what is retained about
the past and whether it is causally deployed.

\noindent\textbf{Memory-augmented methods.}
A crowded line of work adds memory to improve performance
\citep{memoryvla2025,memer2025,torne2025longcontext,contextvla2025,tracevla2024}. These methods are our
\emph{motivation}, not our competition: they presuppose that native VLAs underuse history, and
our diagnosis both confirms that and refines it into an actionable lesson (add unique
information, not more history).

\noindent\textbf{Interpretability tooling.}
We build on linear probing \citep{alain2017probes,belinkov2022probing}, causal-mediation and
interchange/activation-patching interventions
\citep{vig2020causal,geiger2021interchange,meng2022rome,wang2023ioi}, dictionary-learning and
sparse-autoencoder analyses \citep{bricken2023monosemanticity,cunningham2024sparse,templeton2024scaling},
the linear-representation and lens literatures \citep{park2024linear,belrose2023tunedlens}, and
activation steering \citep{turner2023activation}, applied here to the previously unexamined
temporal substrate of action policies. Concurrent VLA-interpretability work observes and controls
features in VLAs \citep{buurmeijer2026observing}, again on the within-timestep axis.

\section{The temporal-deployment audit}
\subsection{Setup}
\label{sec:method}

\noindent\textbf{Models and regime.}
VLAs built on pretrained vision-language backbones now span discrete action-token policies
\citep{rt2_2023,openvla2024,fast2025}, continuous-regression and diffusion/flow action heads
\citep{pi0_2024,pi05_2025,diffusionpolicy2023,cogact2024,openvlaoft2025}, and efficient open
models \citep{smolvla2025,gr00t2025}, trained on large cross-embodiment corpora
\citep{openx2023,droid2024}. All our experiments are on \emph{frozen} public checkpoints, run
\emph{open-loop} (we read action distributions from \texttt{(image, instruction)} without
environment rollout), on a single consumer GPU, with \emph{zero} training. Our primary model is
Octo-Small (27M, window of two frames, diffusion action head) \citep{octo2024}; we add Octo-Base
(93M) for scale and CronusVLA-0.5B (a Qwen-class backbone with a diffusion transformer head,
history carried as a per-frame \emph{cognition} feature) \citep{cronusvla2025} as a second
architecture family. Stimuli are in-distribution Bridge \citep{bridgev2_2023} tuples
$(\mathrm{obs}_{t-1},\mathrm{obs}_t,\text{instruction})$.

\noindent\textbf{Markov vs.\ non-Markov.}
The policy is a window-2 conditional $\pi(a_t \mid o_{t-1}, o_t, c)$ over the previous and current
observations $o_{t-1},o_t$ and instruction $c$, with $s_t$ the underlying scene state.
Behavior-cloned demonstrations are \emph{near-Markov}: the expert action depends almost only on the
present, $a_t^\star\!\approx\!f(s_t)$, so $o_t$ is a near-sufficient statistic and the history
$o_{t-1}$ is redundant. A step is \emph{non-Markov} when $o_t$ underdetermines $a_t^\star$ and
$o_{t-1}$ carries decision-relevant information not recoverable from $o_t$ (i.e.\ $a_t^\star =
f(s_t,s_{t-1})$ with the $s_{t-1}$ dependence not reconstructable from $o_t$). We induce controlled
non-Markovness by corrupting only the current frame: the \texttt{markov} condition leaves $o_t$
intact (present sufficient, history redundant), while \texttt{occ\_black} destroys $o_t$ (forcing
any competent policy onto $o_{t-1}$). The deploy gap $\Delta_\ell$ (Eq.~\ref{eq:gap}) is exactly the
extra causal reliance on history created by that induced non-Markovness, so a policy that fails to
deploy history even here treats the past as redundant by construction.

\noindent\textbf{Frozen stimulus set.}
We build a frozen, content-hashed set of $250$ matched pairs (hash
\texttt{bb4992ac8bbc6803}, $213$ unique instructions), recorded before any analysis to prevent
post-hoc curation. The instruction is held fixed within each matched cell so that any change in
the action is attributable to history rather than to the instruction.

\noindent\textbf{Readouts.}
The action is the mean over $N$ diffusion samples of the predicted $7$-DoF action chunk,
de-normalized via the policy's dataset statistics; we report sample dispersion alongside the
mean. We write $\Delta\!\,\mathrm{action}(A,B)$ for the mean per-dimension $L_2$ distance
between two de-normalized actions. Two derived quantities anchor the study:
\emph{history-decode}@$\ell$ (ridge-probe $R^2$ of the $t\!-\!1$ scene from layer-$\ell$
history-token activations) and \emph{history-contribution}@$\ell$ (the noise-corrected action
shift from a layer-$\ell$ interchange, normalized so that $0$ is no deployment and $1$ is full
deployment). The gap between this contribution under occlusion and under the clean condition is
the deployment signal we track throughout; a large positive gap at a layer where decoding is
high is the signature of present-but-not-remembered.

\subsection{Measurements and audit procedure}
\noindent\textbf{Decode leg: what is present versus what is unique.}
We separate two questions: is the past \emph{present} in the representation, and is any of it
\emph{unique} to the past? Let $\phi^{(\ell)}(x)$ be the pooled history-token activations at layer
$\ell$ and $s_{t-1}$ the previous-frame scene state. A ridge probe $g_\ell$ regresses $s_{t-1}$ on
$\phi^{(\ell)}$, and we report its cross-validated coefficient of determination,
\begin{equation}
A_1(\ell) \;=\; R^2\!\big(g_\ell(\phi^{(\ell)}),\; s_{t-1}\big),
\label{eq:a1}
\end{equation}
which answers \emph{present}. To ask what is \emph{unique}, we project the target onto what the
current frame already explains ($\hat{s}_{t-1\mid t}$, a linear read of $s_{t-1}$ from a
current-frame probe) and decode only the residual,
\begin{equation}
A_4(\ell) \;=\; R^2\!\big(g_\ell(\phi^{(\ell)}),\; s_{t-1}-\hat{s}_{t-1\mid t}\big).
\label{eq:a4}
\end{equation}
$A_4$ is the load-bearing quantity: a large $A_1$ with $A_4\!\approx\!0$ means the history tokens
merely re-encode the present. We sweep $81$ probe configurations (layer, pooling, target,
regularization) to establish the \emph{ceiling} of $A_4$, and use a shuffled-label control to
rule out overfitting.

\noindent\textbf{Deploy leg: noise-corrected interchange.}
Presence in the representation need not mean use in the action; the deploy leg measures use
directly. At layer $\ell$ we overwrite the history-token activations of $x$ with those of a donor
episode $d$ and read the change in the diffusion-averaged action $a(\cdot)$,
\begin{equation}
C^{\mathrm{raw}}_\ell(x,d) \;=\; \big\lVert a\big(\mathrm{patch}_\ell(x;d)\big)-a(x)\big\rVert .
\label{eq:craw}
\end{equation}
Because the diffusion head is stochastic, a self-swap ($d\!=\!x$, whose true contribution is
zero) leaves a non-zero null floor $\approx\!0.15$ (Figure~\ref{fig:noise}); we subtract it to
obtain the corrected contribution $C_\ell = C^{\mathrm{raw}}_\ell-C^{\mathrm{null}}_\ell$ with
$C^{\mathrm{null}}_\ell(x)=C^{\mathrm{raw}}_\ell(x,x)$. The conditional signal is the \emph{deploy
gap} between a current-frame-degraded condition and the clean \texttt{markov} condition,
\begin{equation}
\Delta_\ell \;=\; C_\ell^{\,\mathrm{occ\_black}}-C_\ell^{\,\mathrm{markov}},
\label{eq:gap}
\end{equation}
reported with paired bootstrap $95\%$ CIs. History is \emph{deployed} where $\Delta_\ell$ excludes
zero (the early band) and its dependence is \emph{severed} where $\Delta_\ell\!\approx\!0$ (from
L6): the dissociation is exactly $A_1$ high while $\Delta_\ell\!\to\!0$.

\noindent\textbf{Degradation ladder.}
Rather than a single occlusion, we read deployment as a dose-response over a graded ladder:
partial occlusion (object region in $t$ hidden but visible in $t\!-\!1$), heavy occlusion,
Gaussian blur, and full \texttt{occ\_black} (current frame entirely blacked, forcing reliance on
$t\!-\!1$). A temporally-shuffled cell (frame order swapped) serves as an order-blindness
control. The ladder lets us distinguish a genuine conditional deployment from an
artifact bound to one extreme stimulus.

\noindent\textbf{Second, orthogonal method: attention knockout.}
To confirm the deploy band without relying on representation interchange, we additionally mask
the readout tokens' attention to the history-token keys at each layer and measure the resulting
action shift. This is a mechanistically distinct intervention (removing access rather than
substituting content); agreement between the two methods on the deploy band is strong evidence
that the localization is not an artifact of either.

\noindent\textbf{Statistical protocol.}
Headline effects are hardened three ways: paired bootstrap CIs, multi-seed sign-consistency
(5 seeds), and a sign-flip permutation test (5000 resamples) on pooled per-pair gaps. We report
an effect as deployed only where all three agree.

\noindent\textbf{Why open-loop.}
Open-loop reading is what makes layer-resolved interchange possible: we can intervene at each
block and observe the action shift, a causal localization that closed-loop rollout cannot
deliver. The cost is that the unit of evidence is a controlled pair rather than an episode; we
compensate with multi-seed replication, a permutation test, and bootstrap intervals, and we
treat scale as orthogonal to the localization claim.

\noindent\textbf{Putting it together: a temporal-deployment audit.}
The four measurements compose into one reusable, training-free procedure
(Algorithm~\ref{alg:tda}) that takes any frozen multi-frame VLA and a matched stimulus set and
returns a \emph{history-deployment profile}: whether the past is present ($A_1$), whether any of
it is unique ($A_4$), where and whether it is deployed ($\Delta_\ell$ and the cutoff
$\ell^\star$), the deployment regime ($\sigma$: fallback vs standing), and whether the pre-cutoff
pathway actually carries injected history into the action (the injectability gate,
Appendix~\ref{app:inject}). The audit needs no weight updates and runs open-loop on a single GPU;
we use it below to profile three models and to causally validate the redundancy result.

\begin{algorithm}[t]
\small
\caption{Temporal-Deployment Audit (TDA) of a frozen VLA}
\label{alg:tda}
\begin{algorithmic}[1]
\Require frozen VLA $\pi$; matched pairs $\mathcal{S}=\{(o_{t-1},o_t,\text{instr})\}$; layers $\mathcal{L}$; occlusion $\mathrm{occ}$
\Ensure profile $(A_1,A_4,\{\Delta_\ell\},\ell^\star,\sigma,\textsc{content-bearing})$
\For{$\ell\in\mathcal{L}$}
  \State $A_1(\ell),A_4(\ell)\gets R^2$ of $\phi^{(\ell)}\!\to\!s_{t-1}$ and $\to(s_{t-1}\!-\!\hat s_{t-1\mid t})$ \Comment{present / unique (Eq.~\ref{eq:a1}--\ref{eq:a4})}
  \State $\Delta_\ell\gets (C^{\mathrm{raw}}\!-\!C^{\mathrm{null}})^{\mathrm{occ}}_\ell-(C^{\mathrm{raw}}\!-\!C^{\mathrm{null}})^{\mathrm{markov}}_\ell$, paired CI \Comment{deploy gap (Eq.~\ref{eq:craw}--\ref{eq:gap})}
\EndFor
\State $\ell^\star\gets$ first $\ell$ with $\Delta_\ell$ CI $\ni 0$;\; $\sigma\gets\operatorname{sign}\Delta_\ell$ at early band \Comment{cutoff; fallback $(+)$ / standing $(-)$}
\State inject own clean $o_{t-1}$ at $\ell\!<\!\ell^\star$ under $\mathrm{occ}$;\; $r\gets\lVert a_{\mathrm{occ}}\!-\!a_{\mathrm{full}}\rVert-\lVert a_{\mathrm{inj}}\!-\!a_{\mathrm{full}}\rVert$ \Comment{injectability gate}
\State $\textsc{content-bearing}\gets(r\!>\!0$, CI $\not\ni 0)$;\; \Return $(A_1,A_4,\{\Delta_\ell\},\ell^\star,\sigma,\textsc{content-bearing})$
\end{algorithmic}
\end{algorithm}

\section{Results}
\label{sec:results}

\subsection{Present, but a redundant copy of the current frame}
\label{sec:decode}
\begin{figure}[t]
\centering
\begin{subfigure}[b]{0.49\linewidth}\centering\includegraphics[width=\linewidth]{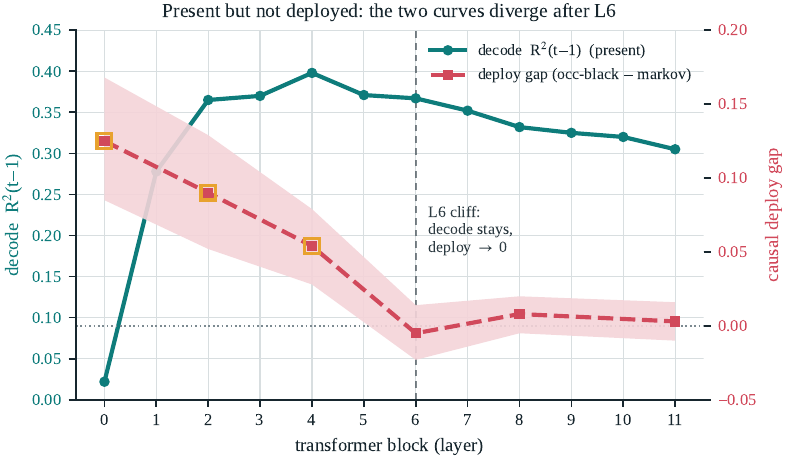}\caption{Present, not deployed: decode stays, gap dies at L6.}\label{fig:dissociation}\end{subfigure}\hfill
\begin{subfigure}[b]{0.49\linewidth}\centering\includegraphics[width=\linewidth]{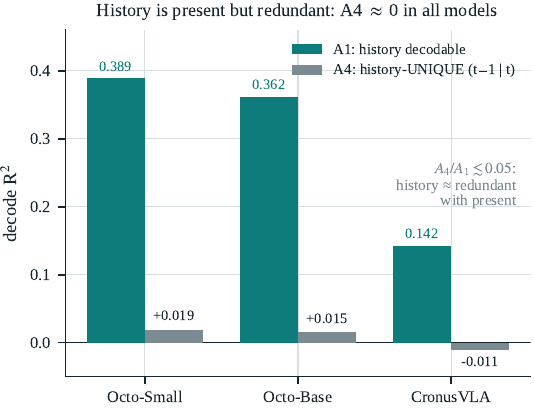}\caption{$A_1$ vs.\ $A_4$: present but redundant.}\label{fig:redundancy}\end{subfigure}

\vspace{4pt}
\begin{subfigure}[b]{0.49\linewidth}\centering\includegraphics[width=\linewidth]{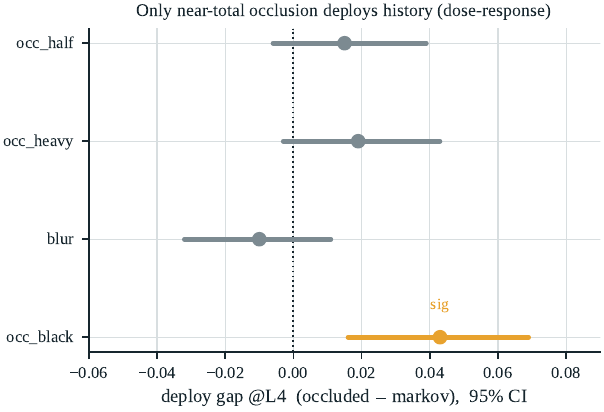}\caption{Deployed only as a fallback (dose-response).}\label{fig:dose}\end{subfigure}\hfill
\begin{subfigure}[b]{0.49\linewidth}\centering\includegraphics[width=\linewidth]{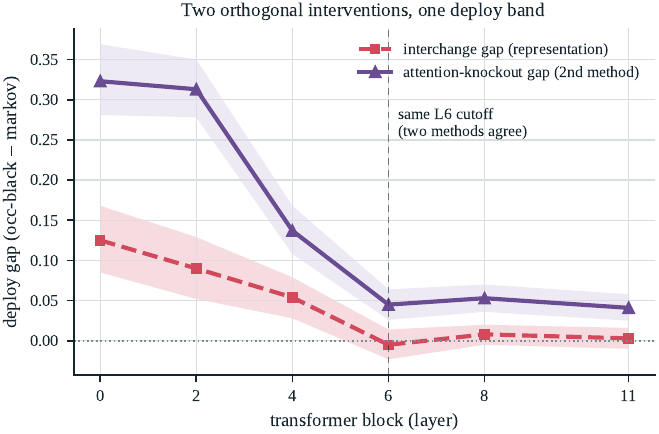}\caption{Two interventions collapse at L6.}\label{fig:twomethods}\end{subfigure}

\vspace{4pt}
\begin{subfigure}[b]{0.49\linewidth}\centering\includegraphics[width=\linewidth]{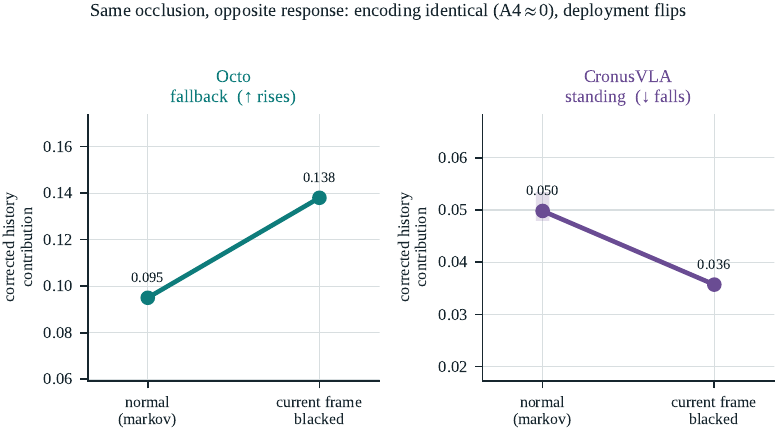}\caption{Same occlusion, opposite response.}\label{fig:signflip}\end{subfigure}\hfill
\begin{subfigure}[b]{0.49\linewidth}\centering\includegraphics[width=\linewidth]{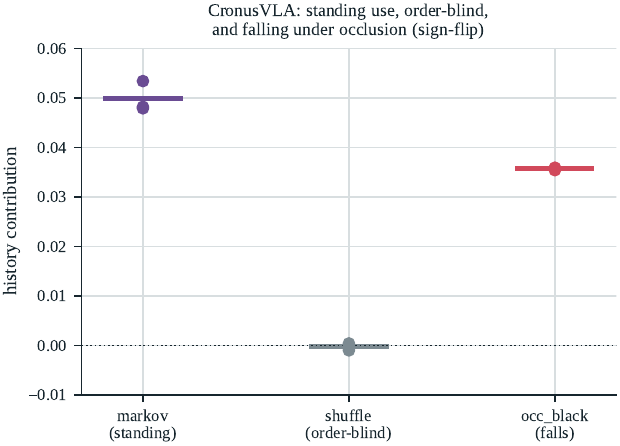}\caption{CronusVLA robustness (standing use).}\label{fig:cronus}\end{subfigure}
\caption{\textbf{The temporal-deployment audit at a glance} (Octo-Small unless noted~\citep{octo2024},
$n{=}200$). Rows group the story: encoding and the central dissociation (a,b), fallback deployment
(c,d), cross-architecture (e,f). \textbf{(a)}~Decode $R^2(t\!-\!1)$ (\present{teal}) stays above
$0.30$ at every depth while the causal deploy gap (\deployc{rose}, occ\_black $-$ markov, $95\%$
band) is significant only through L4 and \emph{collapses at L6}. \textbf{(b)}~History is decodable
($A_1$) but its \emph{unique} information ($A_4$, \ns{gray}) is $\approx\!0$ in every model and both
families: a redundant copy of the present. \textbf{(c)}~On the degradation ladder only near-total
occlusion (\sig{occ\_black}) deploys history; milder degradations are n.s. \textbf{(d)}~Interchange
(\deployc{rose}) and attention-knockout (\textcolor{pnrCronus}{plum}) both collapse at L6, ruling
out a method artifact. \textbf{(e)}~Under the \emph{same} occlusion Octo's reliance on history
\emph{rises} (fallback) while CronusVLA's \emph{falls} (standing), encoding identical
($A_4\!\approx\!0$ in both). \textbf{(f)}~CronusVLA's contribution is positive when clean, $\approx\!0$
under temporal shuffling (order-blind), and falls under occlusion, over three seeds.}
\label{fig:main}
\end{figure}

Past-frame content is linearly decodable from Octo-Small with a peak of $R^2\!=\!0.398$ at L4,
rising from $\approx\!0$ at the input edge (L0) and remaining above $0.30$ through the final
block (Figures~\ref{fig:dissociation} and~\ref{fig:decode_layers}, teal). By the \emph{present} criterion alone the model
clearly registers history, at every depth past the first layer; the decode peak (L4) coincides
with the layer that anchors the causal deploy gap (Section~\ref{sec:deploy}), so history is
\emph{most} decodable and \emph{most} deployed at one layer, after which decoding persists but
deployment is severed. The margin over a
current-frame-only baseline is small ($+0.014$ to $+0.025$ per layer), which already hints at
the redundancy we quantify next.

\noindent\textbf{But redundant: history carries almost no unique information.}\label{sec:redundancy}
The decodable history is overwhelmingly a copy of the present. The history-\emph{unique}
quantity $A_4$ tops out at $+0.019$ for Octo-Small across all $81$ probe configurations, not a
fragile single number but a ceiling, an order of magnitude below the raw decode
($A_4/A_1\!=\!0.019/0.389\!\approx\!0.05$): the decodable history is almost entirely redundant
with the current frame.\footnote{We read $A_4/A_1$ as a heuristic redundancy ratio of two
$R^2$ point estimates, not a formal variance decomposition; the qualitative claim ($A_4\!\approx\!0$
while $A_1$ is large) does not depend on the ratio.} The same holds across models and
architectures: $A_4\!=\!+0.019$ (Octo-Small), $+0.015$ (Octo-Base), and $-0.011$ (CronusVLA),
against $A_1\!=\!0.389/0.362/0.142$ respectively (Figure~\ref{fig:redundancy},
Table~\ref{tab:crossarch}); for CronusVLA, whose history lives in a different representational
locus (a per-frame cognition feature), $A_1$ is read cautiously and $A_4$ is the comparable
quantity (Appendix~\ref{app:cronusfix}). A shuffled-label control returns to $\approx\!0$,
confirming the probe is healthy rather than overfitting. This pre-empts the natural objection
that a decode of $0.39$ merely reflects $t\!-\!1\!\approx\!t$: yes, and we measured exactly how
much (almost all of it).

\noindent\textbf{Scope of the redundancy claim.}
Since $A_4$ (Eq.~\ref{eq:a4}) residualizes against the raw $t\!-\!1$ frame, the one history
quantity with controllable ground truth, we cannot exclude that a policy stores some
\emph{abstract} history (task progress, intent) that is not a linear read of the previous frame.
Two things bound this. First, $A_4\!\approx\!0$ is a ceiling over $81$ probe configurations built
to maximize it, so within the linear-frame hypothesis the redundancy is not an unlucky-probe
artifact. Second, the central \emph{present-but-not-deployed} conclusion does not rest on the
decode target at all: the deploy legs measure the causal effect on the action directly, agnostic
to what the activations encode, and that effect also vanishes by L6. $A_4$ explains \emph{why}
(little unique frame content); the deploy legs establish \emph{that} independently.

\subsection{Not remembered: deployed only as a fallback}
\label{sec:deploy}
Causally, history is read into the action only when the present fails. On the degradation
ladder, only full occlusion (\texttt{occ\_black}) produces a significant deploy gap at L4
($+0.043$, $95\%$ CI $[0.016,0.069]$); partial occlusion, heavy occlusion, and blur are all
not significant (Figure~\ref{fig:dose}). On the unified $n\!=\!200$ source that also drives
Figure~\ref{fig:dissociation}, the gap is significant through L4 ($+0.054$ at L4, CI
$[0.028,0.079]$) and \emph{collapses at L6} (Table~\ref{tab:deploy}): the readout severs its
dependence on history by the middle of the network even though decoding stays high
there.\footnote{The occ\_black gap@L4 varies slightly across sub-studies (different $n$ and donor
pools): $+0.054$ ($n{=}200$ main, Table~\ref{tab:deploy}), $+0.043$ ($n{=}160$ ladder,
Table~\ref{tab:dose_full}), $+0.051$ (Gate-0, Table~\ref{tab:gate_deploy}); all CIs exclude zero
and agree on the L6 collapse.} The
gap survives multi-seed replication and a permutation test ($p\!=\!0.0004$ at L4, significant
in $5/5$ seeds), and an orthogonal attention-knockout intervention recovers the \emph{same} L6
cutoff, with only a small residual past L6. Ecologically, the non-Markov steps that would force history use (Section~\ref{sec:method}) are
rare: a model-free scan of the frozen set finds naturally occluded pairs in only $3$ of $250$ cases, so
the fallback is seldom triggered in distribution (Section~\ref{sec:population}). We next confirm
the L6 cutoff with a second, orthogonal intervention (Section~\ref{sec:twomethods}), harden it
statistically (Section~\ref{sec:harden}), and characterize its scaling
(Section~\ref{sec:scale}). The per-layer contribution under both conditions
(Figure~\ref{fig:contriblayers}) makes the closure visible: the conditional gap (amber band) is
wide early and pinches shut at L6.

\begin{table}[t]
\centering
\caption{\textbf{Layer-resolved causal deployment of history (Octo-Small~\citep{octo2024}, $n{=}200$).}
Corrected interchange contribution under the \texttt{markov} and \texttt{occ\_black}
conditions, and their paired gap with bootstrap 95\% CIs. The gap is significant only
through L4 and \emph{collapses at L6}, while decoding stays high at every depth
(Fig.~\ref{fig:dissociation}): history is \present{present} but not \deployc{deployed}.}
\label{tab:deploy}
\small
\setlength{\tabcolsep}{6pt}
\rowcolors{2}{pnrPaper}{white}
\begin{tabular}{c c c c c c}
\toprule
\textbf{layer} & \textbf{$c_{\text{markov}}$} & \textbf{$c_{\text{occ\_black}}$} & \textbf{gap} & \textbf{95\% CI} & \textbf{sig.}\\
\midrule
\rowcolor{pnrAccentT} 0  & 0.356 & 0.481 & $+0.125$ & $[0.085,\,0.168]$ & \sig{yes}\\
\rowcolor{pnrAccentT} 2  & 0.271 & 0.361 & $+0.090$ & $[0.052,\,0.129]$ & \sig{yes}\\
\rowcolor{pnrAccentT} 4  & 0.087 & 0.141 & $+0.054$ & $[0.028,\,0.079]$ & \sig{yes}\\
6  & 0.045 & 0.040 & $-0.005$ & $[-0.023,\,0.014]$ & \ns{no}\\
8  & 0.019 & 0.027 & $+0.008$ & $[-0.005,\,0.020]$ & \ns{no}\\
11 & 0.000 & 0.000 & $+0.003$ & $[-0.010,\,0.016]$ & \ns{no}\\
\bottomrule
\end{tabular}
\end{table}

\begin{table}[t]
\centering
\caption{\textbf{Cross-architecture summary: same encoding, opposite deployment, opposite steerability} (Octo~\citep{octo2024}, CronusVLA~\citep{cronusvla2025})\textbf{.}
History is linearly decodable (A1) but carries almost no information \emph{unique} to the
past (A4~$\approx 0$) in every model and \emph{both} architecture families; the architectures
differ in \emph{when} history is read into the action (fallback vs.\ standing) and in \emph{whether}
injected history steers the action (Steer: paired donor$-$random directional steer,
Appendix~\ref{app:inject}). Octo's readout is content-blind across three battlefields (G3/G3b/G4;
\emph{inert}); CronusVLA's standing channel is steerable (S3). Octo-Base steer
not separately tested.}
\label{tab:crossarch}
\small
\setlength{\tabcolsep}{6pt}
\rowcolors{2}{pnrPaper}{white}
\begin{tabular}{l c c l c}
\toprule
\textbf{model} & \textbf{$A_1$ (decodable)} & \textbf{$A_4$ (unique)} & \textbf{deployment} & \textbf{steer}\\
\midrule
Octo-Small (27M)~\citep{octo2024} & \present{0.389} & \ns{$+0.019$} & fallback & \ns{inert}\\
Octo-Base (93M)~\citep{octo2024} & \present{0.362} & \ns{$+0.015$} & fallback (replicates) & \ns{--}\\
CronusVLA-0.5B~\citep{cronusvla2025} & \present{0.142} & \ns{$-0.011$} & \textcolor{pnrCronus}{standing} & \textcolor{pnrCronus}{$+0.167$}\\
\bottomrule
\end{tabular}
\end{table}

\noindent\textbf{A second, orthogonal method confirms the L6 cutoff.}\label{sec:twomethods}
Interchange substitutes the \emph{content} of history; attention-knockout removes the readout's
\emph{access} to it. The two interventions are mechanistically distinct, yet they agree on the
deploy band (Figure~\ref{fig:twomethods}, Table~\ref{tab:knockout}): the knockout gap is large
in the early layers (L0/L2 $\approx\!+0.32$), large at L4 ($+0.137$), and drops roughly
sevenfold at L6 ($+0.045$), the same cutoff interchange finds. Knockout leaves a small but
significant residual past L6 (interchange falls to a statistical zero), which is expected and
interpretable: under full occlusion the readout still perturbs whenever any history key is
removed. That the divider is L6 under both methods is what matters (per-layer gaps in
Table~\ref{tab:knockout}).

\begin{table}[t]
\centering
\caption{\textbf{Attention-knockout gap by layer} (Octo-Small~\citep{octo2024},
readout$\rightarrow$history, all heads; gap $=$ occ\_black $-$ markov $\Delta$). An
orthogonal intervention recovers the same L6 cutoff as interchange, with a small significant
residual past L6 (under full occlusion the readout still perturbs when any history key is removed).}
\label{tab:knockout}
\small
\setlength{\tabcolsep}{5pt}
\rowcolors{2}{pnrPaper}{white}
\begin{tabular}{l c c c c c c}
\toprule
\textbf{layer} & \textbf{L0} & \textbf{L2} & \textbf{L4} & \textbf{L6} & \textbf{L8} & \textbf{L11}\\
\midrule
gap & $+0.323$ & $+0.313$ & $+0.137$ & $+0.045$ & $+0.053$ & $+0.041$\\
95\% CI & {\scriptsize[.28,.37]} & {\scriptsize[.28,.35]} & {\scriptsize[.11,.17]} & {\scriptsize[.03,.06]} & {\scriptsize[.04,.07]} & {\scriptsize[.03,.06]}\\
\bottomrule
\end{tabular}
\end{table}

\noindent\textbf{The deployment gap is statistically hardened.}\label{sec:harden}
The early-band gap survives all three hardenings (Table~\ref{tab:harden}, Figure~\ref{fig:hardening}). Across five seeds the
L0 and L4 gaps are positive in $5/5$ seeds; the permutation test gives $p\!=\!0.0000$ (L0) and
$p\!=\!0.0004$ (L4). At L6 and L8 the gap is permutation-indistinguishable from zero
($p\!=\!0.49,\,0.34$): the readout-band non-deployment is a statistical zero, not merely a small
number. L4 is the clean mechanistic anchor (L0 sits at the input edge and may partly reflect
un-mixed input). Three independent significance routes, bootstrap CI, multi-seed
sign-consistency, and permutation, agree.

\subsection{Scaling and cross-architecture}
\label{sec:scale}
\noindent\textbf{Inverse scaling.} Octo-Base ($93$M) reproduces the entire pattern (history deployed only under full occlusion,
gap significant at L0, severed by L6) but at a markedly lower rate. After normalizing each
model by its own within-model full-swap denominators (to control for the wider model's smaller
relative perturbation), Octo-Base deploys history at $\approx\!31\%$ of Octo-Small's rate, with
the same ratio under two independent normalizers (Figure~\ref{fig:inverse},
Table~\ref{tab:inverse}). The $\sim\!11\times$ absolute drop is therefore only partly dilution:
the larger policy relies on history \emph{proportionally} less. If anything, this strengthens
the diagnosis at scale.

\noindent\textbf{Cross-architecture: encoding identical, deployment flips.}\label{sec:crossarch}
The sharpest result is a single intervention with opposite signs. Both architectures encode
$\approx\!0$ unique history ($A_4$ above), yet under the same current-frame occlusion their
reliance on history moves in opposite directions (Figure~\ref{fig:signflip}). Octo's history
contribution \emph{rises} under occlusion, history is an emergency backup invoked only when the
present fails (\emph{fallback}). CronusVLA's contribution is high in the normal condition
($+0.0498$, CI $[0.0479,0.0534]$, $3/3$ seeds) and \emph{falls} under occlusion ($+0.0357$, CI
$[0.0354,0.0359]$; the intervals do not overlap), history is a routine, standing input that is
not escalated by crisis (\emph{standing use}). Frame-order shuffling drives the contribution to
$\approx\!0$ in both architectures, so order-blindness is a shared invariant. The architecture
difference is therefore purely in \emph{when} history is read, not in \emph{how much} unique
history is stored, a dissociation that the encoding alone could never reveal.

Because the two architectures operate at different contribution scales (CronusVLA's
single-locus cognition feature yields smaller absolute shifts than Octo's token band), we read
the sign-flip \emph{within} each architecture, the \emph{direction} of the markov$\to$occlusion
change relative to that model's own baseline, rather than comparing absolute magnitudes across
families. The claim is thus robust to the scale mismatch: Octo's ratio is $>\!1$ (rises),
CronusVLA's is $<\!1$ (falls), and each is established against its own seeds and bootstrap CIs
(Appendix~\ref{app:extra}, Table~\ref{tab:cronus_harden}).

\subsection{Population: how often is the fallback even triggered?}
\label{sec:population}
A fallback matters only if the triggering condition occurs. A model-free scan of the frozen set
for naturally occluded pairs (large inter-frame difference together with object loss or
darkening) finds only $3$ of $250$ candidates at the loosest threshold and none at stricter
ones; the $90$th-percentile lost-fraction is just $0.055$ (Figure~\ref{fig:population_fig}). Genuinely non-Markov situations, in the $o_t$-insufficient sense of
Section~\ref{sec:method}, are thus rare in distribution, consistent with a policy that can afford to treat history as a
last-resort backup, and with the observation that adding memory helps mainly on the curated,
history-dependent tasks that long-horizon benchmarks such as LIBERO~\citep{libero2023} construct.
Behaviorally, the small history dependence that does exist is not confined to one output channel:
swapping $t\!-\!1$ perturbs every continuous pose dimension, and the gripper accounts for only
$\approx\!9\%$ of the summed effect (Figure~\ref{fig:perdof}).

\subsection{Synthesis and implications}
\label{sec:discussion}
The three layers interlock into one statement: \emph{for a frozen VLA, history is not an
independently encoded working memory but a redundant residual of the current frame, deployed
into the action only as a fallback when the present fails.} This reframes the common claim that
\emph{VLAs are memoryless.} They are not memory-incapable; they store a copy of the present labeled
as the past, and they read it only under duress. That is why such policies fail on genuinely
non-Markov tasks: not because history cannot be stored, but because what is stored duplicates
the very frame that has been lost.

The design implication is direct: because the native substrate already encodes the
present-as-history, adding more history (longer windows, richer overlays) adds mostly redundancy;
the payoff is in mechanisms that inject information \emph{unique} to the past and make the readout
attend to it before the mid-network cutoff where the native dependence is severed.

\noindent\textbf{Causal validation, and an architecture-conditional steerability flip.}
The $A_4\!\approx\!0$ result \emph{predicts} that in the fallback regime the stored history, a copy
of the present, is causally inert: re-supplying it cannot add what was never encoded. The audit's
injectability gate confirms this on Octo (Appendix~\ref{app:inject}): injection at a pre-cutoff
layer is localized and donor-specific, yet across three independent battlefields it is
sub-behavioral, it neither repairs occlusion, nor disambiguates state-aliasing, nor opens the
fallback gate via the (perfectly decodable, probe acc $1.0$) temporal-identity direction. Octo's
pathway is \emph{content-blind}: it registers that \emph{something} was injected, not \emph{what}.
This is not universal. On the standing-use architecture, whose history channel feeds the action
continuously rather than being severed at a cutoff, the \emph{same} injection \emph{does} steer the
action toward the donor (paired steer\,$-$\,random $+0.167$, CI\,$[0.106,0.227]$; a
reliable directional effect). Content-steerability therefore tracks the
\emph{deployment regime}, a third architecture-conditional dissociation on the same axis (encoding
identical, deployment sign-flipped, steerability sign-flipped), and the reading is causal:
injection is inert exactly where the readout severs history and effective exactly where it does
not. The lesson sharpens: a fallback model's native window cannot be unlocked by injection, so
external memory must add present-irreducible information through an engineered pathway, whereas a
standing-use channel is directly writable.

\begin{table}[t]
\centering
\caption{\textbf{Injectability gate across two architectures} (frozen, open-loop, training-free;
injection at a pre-cutoff layer, L4 for Octo). In the \emph{fallback} regime (Octo) injection is
localized and donor-specific but \emph{content-blind} on three independent battlefields
(G3/G3b/G4), causally confirming $A_4\!\approx\!0$; in the \emph{standing} regime (CronusVLA) the
\emph{same} injection steers the action (S3). Steerability tracks the deployment regime. Octo:
full $250$-pair set, three seeds; S3: three seeds and three donors on a held-out stimulus subset.}
\label{tab:inject}
\small
\setlength{\tabcolsep}{4pt}
\rowcolors{2}{pnrPaper}{white}
\begin{tabular}{l l l l l}
\toprule
\textbf{check} & \textbf{model (regime)} & \textbf{question} & \textbf{statistic} & \textbf{verdict}\\
\midrule
G0a & Octo (fallback) & injectable pre-cutoff? & L4 contrib.\ $7\text{--}9\times$ L8 & \present{localized}\\
G0b & Octo & donor-specific info? & sep$-$null $+11.0$ [9.6, 12.5] & \present{specific}\\
G2 & Octo & steers to donor action? & pre$-$null $+0.069$ [.043, .094] & \sig{yes, modest}\\
G3 & Octo & repairs occlusion? & recovered $-0.010$ [$-.42$, $+.41$] & \ns{no}\\
G3b & Octo & disambiguates aliasing? & paired $+0.004$ [$-.12$, $+.11$] & \ns{no}\\
G4 & Octo & flips identity gate? & paired $-0.040$ [$-.071$, $-.009$] & \ns{no}\\
S3 & CronusVLA (standing) & steers to donor action? & paired $+0.167$ [.106, .227] & \present{\textbf{yes}}\\
\bottomrule
\end{tabular}
\end{table}

\noindent\textbf{A combined mechanistic picture.}
Read alongside the fusion-axis result of \citet{grant2026notall}, our findings complete a
two-axis description of the VLA action mechanism. On the \emph{modality} axis the mechanism is
shared across architectures (vision dominates, architecture-independently); on the \emph{time}
axis it is shared in \emph{what} is encoded (history-unique information $\approx\!0$ in both
families) but divergent in \emph{when} it is deployed (fallback vs.\ standing). The two studies
are not competing increments but orthogonal coordinates: one localizes space within a timestep,
the other information across timesteps. A reader who accepts Grant's diagnosis should find ours
the necessary other half, and should be surprised by the same-encoding/opposite-deployment
dissociation, which the fusion axis cannot surface.

\noindent\textbf{Why the substrate is redundant.}
The redundancy is not specific to one checkpoint; it recurs across $27$M, $93$M, and a $0.5$B
model spanning two families, and deepens with scale (Section~\ref{sec:scale}). A policy trained to
imitate near-Markov demonstrations is never pressured to store anything about $t\!-\!1$ not already
legible in $t$, so the window-of-two history becomes a near-copy of the current frame and the
readout learns to consult it only when the present is uninformative. This predicts that memory
modules trained on the same data inherit the redundancy unless their objective explicitly rewards
present-irreducible information, a testable hypothesis our diagnosis hands the memory line.

\noindent\textbf{What the result does \emph{not} claim.}
We do not claim VLAs \emph{cannot remember,} nor that memory modules are useless. The claim is
sharper and survives its own strongest objection: history is present as redundancy and deployed
as a fallback, and the decode-vs-deploy and $A_1$-vs-$A_4$ gaps quantify exactly how much. The
precision is the point, an over-broad \emph{memoryless} slogan would be both wrong and
unfalsifiable, whereas the three-layer statement is falsifiable and was, in one of its
sub-hypotheses (\emph{standing use encodes more history}), cleanly falsified by our own data.

\noindent\textbf{Future work.}
Three extensions follow directly: (i)~a \emph{feature-level} $A_4$ via dictionary learning
\citep{bricken2023monosemanticity}, turning the redundancy result from a population statistic into
a per-feature target to amplify; (ii)~a \emph{closed-loop} study measuring whether injecting
present-irreducible information at the readout band improves non-Markov task success; and
(iii)~\emph{more architecture members} (Appendix~\ref{app:survey}) to turn the fallback-vs-standing
dissociation into a taxonomy.

\section{Conclusion}
\label{sec:conclusion}
VLAs do not forget the past so much as they were never built to treat it as separate from the
present. History is present in the representation, but largely as a redundant copy of the current
frame, and it is deployed into the action only as a fallback when the present fails, with the
deployment regime, but not the encoding, depending on architecture. The diagnosis the
memory-augmentation literature presupposes turns out to recommend a specific fix: add what is
unique, not what is more.

\noindent\textbf{Limitations.} Two design choices are deliberate, not concessions.
\emph{Open-loop} evaluation on $\sim\!250$ controlled pairs is what makes layer-resolved causal
localization possible: interchange and attention-knockout at \emph{every} block require a fixed,
repeatable input that closed-loop rollout cannot provide, and a purely behavioral study can inject
activations but cannot read the per-layer severance we localize; a closed-loop, real-robot
extension is complementary future work that does not bear on the localization claim.
\emph{Two architecture families} suffice to \emph{establish} the fallback-vs-standing
dissociation, which needs only the two regime endpoints; a systematic search for a third frozen,
open, multi-frame, consumer-GPU-runnable family returned none (Appendix~\ref{app:survey}), and
further members would \emph{populate} the axis rather than prove it. Finally, a single consumer GPU
bounds model size, though the inverse-scaling tendency (Section~\ref{sec:scale}) suggests larger
policies rely on history even less, strengthening the diagnosis.

\section*{Reproducibility Statement}
All experiments use \emph{frozen}, publicly available checkpoints (Octo-Small/Base~\citep{octo2024},
CronusVLA-0.5B~\citep{cronusvla2025}) with \emph{zero} training, run open-loop on a single
consumer GPU. The stimulus set is a content-hashed, frozen collection of $250$ matched pairs
(hash \texttt{bb4992ac8bbc6803}, $213$ unique instructions) drawn from BridgeData~V2~%
\citep{bridgev2_2023}; the build script plus the hash are released in lieu of the raw corpus.
Every headline number is bound to a specific run and reported with its sample size and paired
bootstrap CI (Appendix~\ref{app:numbers}); the probe, interchange, and attention-knockout
procedures, the self-swap noise calibration, and the pinned software environments (two conda
specifications, exact checkpoint identifiers) are documented in
Appendices~\ref{app:probe}--\ref{app:repro}. The per-leg reproduction map
(Table~\ref{tab:reproduce}) lists each module and its expected headline value, and the harness is
certified by a backbone-parity check (Appendix~\ref{app:harness}). Statistical claims are made
only where bootstrap CI, five-seed sign-consistency, and a permutation test agree
(Section~\ref{sec:harden}).

\section*{Ethics Statement}
This work is a mechanistic analysis of existing, publicly released robot-learning checkpoints and
datasets; it introduces no new human-subjects data and no new model release. The stimuli are
frames from BridgeData~V2, an established academic manipulation dataset, used under its terms and
distributed only as a content hash plus a build script. Because our findings characterize
\emph{when} a policy does or does not rely on visual history, they carry a dual-use consideration:
the same diagnosis that guides safer memory design could inform an adversary constructing
occlusion conditions under which a deployed policy silently falls back on stale history. We judge
this risk low relative to the transparency benefit, since the failure mode (near-total
current-frame loss) is already visible to any operator and the analysis operates open-loop on
frozen weights. We use no proprietary data, report negative and null results faithfully, and
flag the open-loop scope as a limitation rather than overstating real-robot safety implications.

\bibliography{references}
\bibliographystyle{iclr2026_conference}

\appendix
\section{Verified numbers}
\label{app:numbers}
All headline values are drawn from a single verified results log. Decode $R^2(t\!-\!1)$ by
layer (L0--L11): $0.022, 0.278, 0.365, 0.370, \mathbf{0.398}, 0.371, 0.367, 0.352, 0.332,
0.325, 0.320, 0.305$; the per-layer deploy gap, with paired bootstrap $95\%$ CIs, is in
Table~\ref{tab:deploy}. The remaining verified tables, Gate-0 decode/deploy arms, the full
dose-response, statistical hardening, the attention-knockout, the normalized cross-scale, and
the behavioral per-DoF breakdown, are collected below.


\begin{table}[h]
\centering
\caption{Gate-0 decode arm (Octo-Small~\citep{octo2024}, ridge probe, $n{=}200$, 5-fold). Past-frame content is decodable
from L2 with a small margin over a current-frame-only baseline.}
\label{tab:gate_decode}
\small
\rowcolors{2}{pnrPaper}{white}
\begin{tabular}{c c c c}
\toprule
\textbf{layer} & \textbf{$R^2(t\!-\!1)$} & \textbf{$R^2(t)$ baseline} & \textbf{margin}\\
\midrule
0 & 0.022 & 0.027 & $-0.005$\\
2 & 0.362 & 0.347 & $+0.015$\\
4 & 0.388 & 0.374 & $+0.014$\\
6 & 0.338 & 0.342 & $-0.004$\\
\bottomrule
\end{tabular}
\end{table}

\begin{table}[h]
\centering
\caption{Gate-0 deploy arm (Octo-Small~\citep{octo2024}, corrected interchange contribution, $n{=}200$). The gap over
\texttt{markov} is significant for \texttt{occ\_black} but not \texttt{occ\_partial} at L4.}
\label{tab:gate_deploy}
\small
\setlength{\tabcolsep}{5pt}
\rowcolors{2}{pnrPaper}{white}
\begin{tabular}{l c c c c c c}
\toprule
\textbf{cell} & \textbf{L0} & \textbf{L2} & \textbf{L4} & \textbf{L6} & \textbf{L8} & \textbf{L11}\\
\midrule
markov      & 0.344 & 0.254 & 0.088 & 0.051 & 0.015 & 0.002\\
occ\_partial & 0.385 & 0.279 & 0.099 & 0.043 & 0.012 & 0.004\\
occ\_black   & 0.472 & 0.367 & 0.139 & 0.049 & 0.015 & 0.000\\
\midrule
\multicolumn{7}{l}{\footnotesize gap@L4: occ\_partial $+0.010\,[-0.008,0.028]$ \ns{n.s.};\ \
occ\_black $+0.051\,[0.026,0.074]$ \sig{sig.}}\\
\bottomrule
\end{tabular}
\end{table}

\begin{table}[h]
\centering
\caption{Stimulus-strength dose-response (Octo-Small~\citep{octo2024}, gate0b, $n{=}160$). Corrected contribution at L2/L4
and the L4 gap vs \texttt{markov}. Only \texttt{occ\_black} is significant; \texttt{shuffle}
(temporal-order destruction) is reported as the order-blindness control.}
\label{tab:dose_full}
\small
\setlength{\tabcolsep}{6pt}
\rowcolors{2}{pnrPaper}{white}
\begin{tabular}{l c c c c c}
\toprule
\textbf{cell} & \textbf{$c$@L2} & \textbf{$c$@L4} & \textbf{gap@L4} & \textbf{95\% CI} & \textbf{sig.}\\
\midrule
markov     & 0.257 & 0.095 &, & (baseline) &, \\
occ\_half   & 0.334 & 0.110 & $+0.015$ & $[-0.006,0.039]$ & \ns{no}\\
occ\_heavy  & 0.324 & 0.114 & $+0.019$ & $[-0.003,0.043]$ & \ns{no}\\
blur       & 0.310 & 0.085 & $-0.010$ & $[-0.032,0.011]$ & \ns{no}\\
\rowcolor{pnrAccentT} occ\_black  & 0.376 & 0.138 & $+0.043$ & $[0.016,0.069]$ & \sig{yes}\\
shuffle    & 0.000 & 0.000 & $-0.098$ & $[-0.116,-0.080]$ & \ns{order-blind}\\
\bottomrule
\end{tabular}
\end{table}

\begin{table}[h]
\centering
\caption{Statistical hardening (Octo-Small~\citep{octo2024}, 5 seeds $\times$ 60 pairs; permutation, 5000 resamples on
$n{=}300$ pooled per-pair gaps). The deploy gap is sign-consistent across all 5 seeds and
permutation-significant at L0/L4; at L6/L8 it is permutation-indistinguishable from zero.}
\label{tab:harden}
\small
\setlength{\tabcolsep}{6pt}
\rowcolors{2}{pnrPaper}{white}
\begin{tabular}{c c c c c c}
\toprule
\textbf{layer} & \textbf{per-seed gap (s0--s4)} & \textbf{mean} & \textbf{std} & \textbf{+seeds} & \textbf{perm $p$}\\
\midrule
\rowcolor{pnrAccentT} 0 & .142/.119/.040/.096/.141 & $+0.108$ & 0.038 & 5/5 & $0.0000$\\
\rowcolor{pnrAccentT} 4 & .037/.009/.023/.047/.054 & $+0.034$ & 0.016 & 5/5 & $0.0004$\\
6 & $-$.001/.005/.013/$-$.008/$-$.008 & $+0.000$ & 0.008 & 2/5 & $0.4882$\\
8 & .018/.000/.024/$-$.012/$-$.016 & $+0.003$ & 0.016 & 3/5 & $0.3376$\\
\bottomrule
\end{tabular}
\end{table}

\begin{table}[h]
\centering
\caption{Normalized cross-scale (Octo~\citep{octo2024}, inverse scaling; within-model full-swap denominators).
After normalization Octo-Base still deploys history at $\approx 31\%$ of Octo-Small's rate.}
\label{tab:inverse}
\small
\setlength{\tabcolsep}{6pt}
\rowcolors{2}{pnrPaper}{white}
\begin{tabular}{l c c c c}
\toprule
\textbf{model} & \textbf{contrib (markov L0)} & \textbf{norm/hist} & \textbf{norm/cur} & \textbf{abs.\ ratio}\\
\midrule
Octo-Small~\citep{octo2024} (27M) & 0.339 & 0.637 & 0.555 &, \\
Octo-Base~\citep{octo2024} (93M) & 0.029 & 0.199 & 0.172 &, \\
\textbf{base/small} & \textbf{0.09} & \textbf{0.31} & \textbf{0.31} &, \\
\bottomrule
\end{tabular}
\end{table}

\begin{table}[h]
\centering
\caption{Behavioral per-DoF $|\Delta|$ (Octo-Small~\citep{octo2024}, markov cell, $n{=}48\times32$ samples). History affects
the whole 7-DoF action vector, not just the discrete gripper (gripper $=9.3\%$ of summed mask
$\Delta$). Scalar behavioral sensitivity is weak (occ\_black/markov ratio $1.22\times$),
motivating a mechanistic rather than behavioral study.}
\label{tab:perdof}
\small
\setlength{\tabcolsep}{6pt}
\rowcolors{2}{pnrPaper}{white}
\begin{tabular}{l c c c c c c c}
\toprule
\textbf{condition} & \textbf{wx} & \textbf{wy} & \textbf{wz} & \textbf{rx} & \textbf{ry} & \textbf{rz} & \textbf{grip}\\
\midrule
mask $t\!-\!1$ & 0.463 & 0.294 & 0.429 & 0.348 & 0.365 & 0.376 & 0.233\\
swap $t\!-\!1$ & 0.720 & 0.477 & 0.538 & 0.409 & 0.392 & 0.553 & 0.260\\
\bottomrule
\end{tabular}
\end{table}

\noindent\textbf{Natural-occlusion scan (population leg).} On the frozen set ($n{=}250$, model-free):
candidates at the loosest bar $=3/250$, at stricter bars $=0$. Frame-difference percentiles
(50/90/95) $=5.08/9.99/12.52$; lost-fraction percentiles $=0.020/0.055/0.070$.

\noindent\textbf{Interchange noise calibration.} Raw donor-swap vs.\ self-swap null (layers
0/4/6/8/11): raw markov $0.666/0.299/0.221/0.177/0.162$, null markov
$0.151/0.150/0.166/0.161/0.158$; the null floor ($\approx\!0.15$) is subtracted to obtain the
corrected contribution. Stimulus hash \texttt{bb4992ac8bbc6803}.

\section{Probe and intervention details}
\label{app:probe}
\textbf{Decode probe.} Ridge regression on history-token activations (mean$+$max pooling, RGB16
frame target), layer-swept, 5-fold; the $A_4$ residual decodes $t\!-\!1$ after projecting out a
current-frame ($t$) probe. The $A_4$ ceiling is established by an $81$-configuration sweep over
\{layer\}$\times$\{pooling\}$\times$\{target\}$\times$\{regularization\}; the best $A_4$ is
$+0.019$ (Octo-Small), $+0.015$ (Octo-Base), and within the standardized CronusVLA arm
$-0.011$. A shuffled-label control returns $A_2\!\approx\!0$, ruling out overfitting.
\textbf{Interchange.} A forward pre-hook on \texttt{encoder\_blocks.}$\ell$ overwrites the $256$
history-token positions (a $16\times16$ patch grid, located empirically and stable across
pairs) with a donor's layer-$\ell$ representation; the current frame and readout are untouched.
\textbf{Token layout.} \texttt{[task lang][$t_0$: primary(256), wrist, readout][$t_1$: primary,
wrist, readout]}, total $690$ tokens at width $384$ (Octo-Small).

\begin{figure}[h]
\centering
\includegraphics[width=0.50\linewidth]{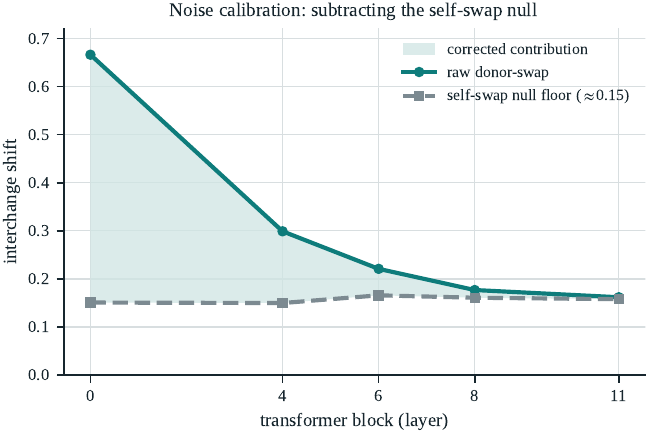}
\caption{Interchange noise calibration (Octo-Small~\citep{octo2024}). The raw donor-swap shift
(teal) minus the self-swap null floor (slate, $\approx\!0.15$, whose true contribution is zero)
gives the corrected contribution (shaded). Every deploy-leg number is this corrected quantity.}
\label{fig:noise}
\end{figure}

\section{Environment and reproducibility}
\label{app:repro}
Experiments use two pinned conda environments. The Octo path pins torch $2.6.0$+cu124,
transformers $4.34.1$ (Flax-capable), a CPU JAX $0.4.20$ used only for weight loading,
tensorflow-cpu $2.15.0$, and numpy $1.26.4$; interpretation uses the PyTorch port (JAX is
weight-loading only). The CronusVLA path pins torch $2.2.0$+cu121, transformers $4.40.1$
(load-bearing), numpy $1.26.4$, tokenizers $0.19.1$, timm $0.9.10$, and \texttt{draccus};
scikit-learn is pinned so it does not upgrade numpy. Models are loaded frozen from public
checkpoints (Octo-Small/Base $1.5$; CronusVLA-$0.5$B Bridge/RT-1 post-trained checkpoint
\texttt{step-042500-epoch-07-loss=0.0587.pt}). The harness is certified by a backbone parity
check (readout mean$|\Delta|=2.4\!\times\!10^{-4}$) and a deterministic readout-parity
tightening. Everything runs on a single $24$GB consumer GPU with no quantization and no
training; the $387$GB Bridge corpus is subsampled to the hashed $250$-pair set, distributed as
the build script plus the hash rather than the raw data.

\section{Third-architecture exclusion survey}
\label{app:survey}
A frozen public checkpoint that is open, history-aware (multi-frame at inference), and runnable
on a consumer GPU is rare. We surveyed the candidate pool and excluded each member for a
principled reason: multi-\emph{view} (not multi-\emph{frame}) fine-tuned policies (e.g.\
OpenVLA-OFT \citep{openvlaoft2025}); flow-matching policies that default to a single frame
(\(\pi_0\)/\(\pi_{0.5}\) \citep{pi0_2024,pi05_2025}, X-VLA \citep{xvla2025}, the CVAE-based ACT
\citep{act2023}); a joint-embedding predictive model whose leakage-free student reduces to a
single frame at inference (its environment was built and weights loaded, $\sim\!6$GB, before
exclusion); and a spatio-temporal RGB-D model with a mandatory depth input absent from our
stimulus set (3D-/4D-VLA \citep{threedvla2024,fourdvla2025}). The two families we
study, Octo-class and Qwen-DiT (CronusVLA), are the rare members meeting all four constraints,
which is why the architecture axis has two points rather than more. We frame this as scope, not
omission: the fallback-vs-standing dissociation is established on the two members that the
constraints admit.

\section{Cross-architecture decode: a standardization fix}
\label{app:cronusfix}
CronusVLA's \emph{history} is a per-frame cognition feature ($896$-d, the VLM's last-layer last
token). A first decode attempt produced $R^2\!\ll\!0$ with the shuffled-label control
\emph{more} negative than the real label ($-0.463$ vs.\ $-0.004$), a numerical breakdown from
ridge-overfitting an unnormalized, large-scale, high-dimensional feature ($p\!\gg\!n$). The
corrected probe z-scores the features, widens the regularization sweep ($10^2$--$10^5$), and
adds a PCA-50 variant; the shuffle control then returns to $\approx\!0$ ($-0.032$) and the
results become trustworthy: $A_1\!=\!0.142$, $A_4\!=\!-0.011$, stable across the full and PCA-50
parameterizations. We report $A_1$ for CronusVLA cautiously (the cognition feature is a
different representational locus than Octo's mid-transformer tokens); the comparable quantity is
$A_4$, and both architectures sit at $\approx\!0$.

\section{Supporting layer-resolved and robustness panels}
\label{app:contriblayers}
\begin{figure}[h]
\centering
\begin{subfigure}[b]{0.49\linewidth}\centering\includegraphics[width=\linewidth]{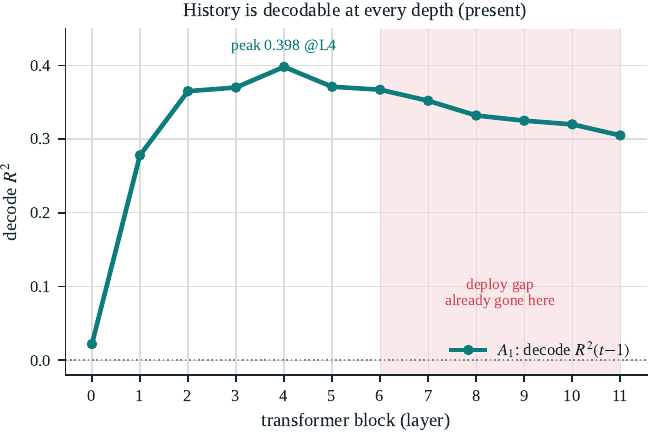}\caption{Decode $A_1$ at every depth.}\label{fig:decode_layers}\end{subfigure}\hfill
\begin{subfigure}[b]{0.49\linewidth}\centering\includegraphics[width=\linewidth]{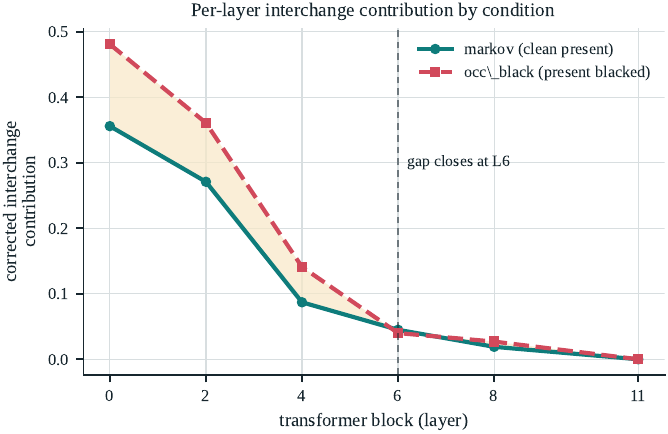}\caption{Per-layer contribution; gap closes at L6.}\label{fig:contriblayers}\end{subfigure}

\vspace{3pt}
\begin{subfigure}[b]{0.49\linewidth}\centering\includegraphics[width=\linewidth]{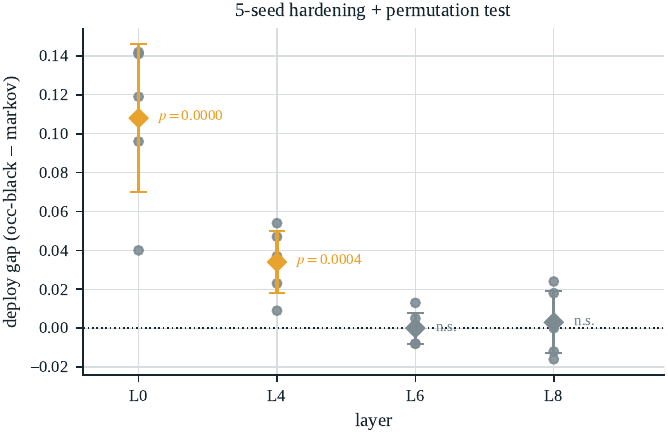}\caption{5-seed hardening $+$ permutation.}\label{fig:hardening}\end{subfigure}\hfill
\begin{subfigure}[b]{0.49\linewidth}\centering\includegraphics[width=\linewidth]{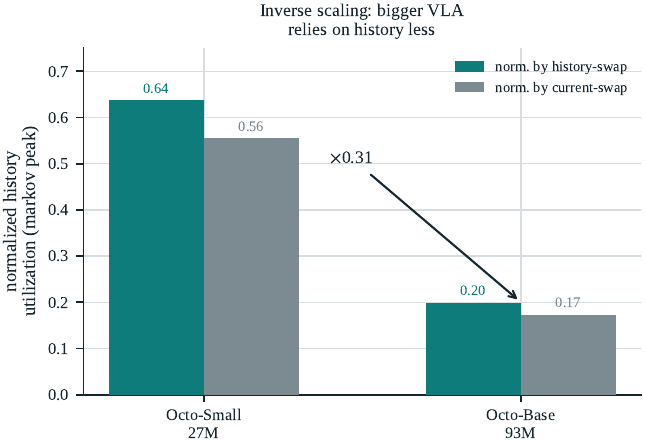}\caption{Inverse scaling across model size.}\label{fig:inverse}\end{subfigure}
\caption{\textbf{Supporting panels} (Octo-Small unless noted~\citep{octo2024}): (a)~decode by
depth, (b)~per-layer interchange contribution, (c)~five-seed deploy-gap hardening with a
permutation test, (d)~inverse scaling across model size. These expand the summary panels of
Figure~\ref{fig:main}.}
\label{fig:appendix_panels}
\end{figure}

Figure~\ref{fig:appendix_panels} collects four panels that the main figure summarizes:
\textbf{(a)}~history-decode $A_1$ across all $12$ layers, decodable at every depth and peaking at
L4; \textbf{(b)}~per-layer corrected interchange contribution under \texttt{markov} and
\texttt{occ\_black}, both riding the same decreasing curve with the conditional gap closing at L6;
\textbf{(c)}~the deploy gap across five seeds with permutation $p$, significant at L0/L4 and a
statistical zero at L6/L8; \textbf{(d)}~Octo-Base deploys history at $\approx\!31\%$ of
Octo-Small's rate under both normalizers. Interchange was evaluated at the canonical layer set
$\{0,2,4,6,8,11\}$ for cost; the 12-layer decode curve shows the representation stays decodable in
the same range where the deploy gap has already vanished.

\begin{figure}[h]
\centering
\begin{subfigure}[b]{0.52\linewidth}
\centering
\includegraphics[width=\linewidth]{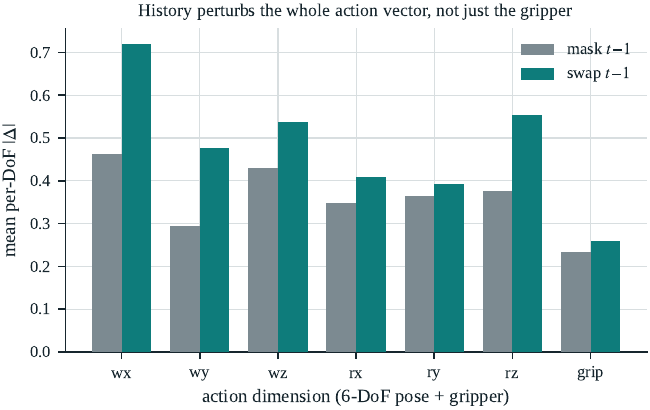}
\caption{Behavioral per-DoF sensitivity.}
\label{fig:perdof}
\end{subfigure}\hfill
\begin{subfigure}[b]{0.46\linewidth}
\centering
\includegraphics[width=\linewidth]{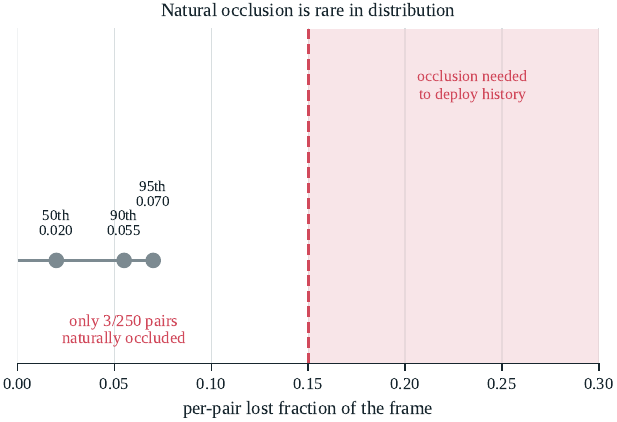}
\caption{Natural occlusion is rare.}
\label{fig:population_fig}
\end{subfigure}
\caption{\textbf{Population and behavioral evidence} (Octo-Small~\citep{octo2024}). (a)~Mean
per-DoF $|\Delta|$ under masking vs.\ swapping $t\!-\!1$: history influences the continuous pose
dimensions throughout, not just the gripper. (b)~Per-pair lost fraction of the frame across the
frozen set; the distribution sits far below the near-total-occlusion band that triggers
deployment, so only $3/250$ pairs are naturally occluded, the fallback is seldom triggered in
distribution.}
\label{fig:perdof_population}
\end{figure}

\section{Injectability gate: can the pre-cutoff pathway carry history content?}
\label{app:inject}
The audit's final leg (Algorithm~\ref{alg:tda}, and the causal validation in
Section~\ref{sec:discussion}) tests whether the pathway that \emph{reads} history can be driven to
\emph{use} injected history content. Six checks across two architectures reuse the noise-corrected
interchange engine (Eq.~\ref{eq:craw}), are training-free, and run open-loop; Octo checks use the
frozen $250$-pair set with three seeds; the CronusVLA check (S3) uses three donors over a held-out stimulus subset. The chain is designed so a \emph{negative} is as informative as a
positive: the fallback-regime negatives are the causal counterpart of $A_4\!\approx\!0$, and the
standing-regime positive shows content-blindness is architecture-conditional, not universal.

\noindent\textbf{Reading.} G0a/G0b/G2 establish that injecting a clean donor's history at a
pre-cutoff layer of Octo is localized (L4 moves the action $7\text{--}9\times$ more than L8),
writes donor-specific information (separation over the self-swap null excludes zero), and imposes a
modest directional bias. The three decisive Octo checks are negative. \textbf{G3} injects the
stimulus's \emph{own} pre-occlusion history under \texttt{occ\_black}: recovery toward the
unoccluded action is a clean zero, while a mismatched injection \emph{harms} (own$-$random $+4.9$,
CI\,$[3.1,6.9]$), so the hook is causally live but the model's own redundant history fills nothing.
\textbf{G3b} moves to state-aliasing (current frame clear but ambiguous): the aliased partner's
history fails to switch the branch once the common drift is removed by a paired contrast, and an
unrelated history moves the action as much. \textbf{G4} pushes the current tokens along the
temporal-identity direction, which a linear probe decodes at accuracy $1.0$; it opens the fallback
gate no more than a norm-matched random push, so \emph{identity}, like \emph{content}, is
decodable but not deployable. Three battlefields agree: Octo's pathway is content-blind and severed
by L6. \textbf{S3} then breaks the universality: on CronusVLA, whose cognition-feature history
channel feeds the action's cross-attention continuously (standing use, no cutoff), injecting a
donor's history steers the action toward that donor (paired $+0.167$; a large non-specific drift remains). The contrast is clean, injection is inert exactly
where the readout severs history and effective exactly where it does not, so we report a causal
\emph{law}, steerability tracks the deployment regime, rather than a control method: a fallback
model's native window cannot be unlocked by injection, whereas a standing-use channel is directly
writable.

\section{Notation, design sweeps, and reproduction}
\label{app:extra}
Table~\ref{tab:notation} fixes notation. Tables~\ref{tab:sweep_small}--\ref{tab:sweep_base} give
the decode design sweep that establishes the $A_4$ ceiling at two scales; Table~\ref{tab:m0b}
gives the behavioral history-sensitivity that motivates a mechanistic measure;
Table~\ref{tab:reproduce} maps each leg to its module and expected headline number; and
Table~\ref{tab:cronus_harden} reports the CronusVLA hardening (standing use robust over seeds,
order-blind, and falling rather than rising under occlusion; Figure~\ref{fig:cronus}). Figure~\ref{fig:perdof} shows that
history perturbs the whole $7$-DoF action vector, not just the discrete gripper, so the deploy
signal is not an artifact of a single output dimension.


\begin{table}[h]
\centering
\caption{Notation.}
\label{tab:notation}
\small
\begin{tabular}{l l}
\toprule
symbol & meaning\\
\midrule
$t,\,t\!-\!1$ & current / previous timestep (observation)\\
$\ell$ & transformer block index (L0--L11 for Octo-Small)\\
$A_1$ & decode $R^2$ of $t\!-\!1$ from history-token activations (\emph{is history present?})\\
$A_4$ & decode of $t\!-\!1$ residualized against $t$ (\emph{history-\emph{unique} info})\\
$A_2$ & shuffled-label decode control ($\approx\!0$ when the probe is healthy)\\
$c@\ell$ & noise-corrected interchange contribution at layer $\ell$\\
gap & $c_{\text{occ\_black}}-c_{\text{markov}}$ (conditional deployment signal)\\
markov & clean current frame (history nominally unnecessary)\\
occ\_black & current frame fully blacked (forces reliance on $t\!-\!1$)\\
shuffle & temporal order destroyed (order-blindness control)\\
\bottomrule
\end{tabular}
\end{table}

\begin{table}[h]
\centering
\caption{Decode design sweep, Octo-Small~\citep{octo2024} (top-8 by $A_4$ over $81$ configs). The
history-\emph{unique} ceiling is $A_4\!=\!+0.019$; the best raw decode is $A_1\!=\!0.389$
(meanmax, RGB16, L2, $\alpha{=}10$, not in this $A_4$-sorted view).}
\label{tab:sweep_small}
\small
\setlength{\tabcolsep}{6pt}
\begin{tabular}{c l l c c c}
\toprule
layer & pool & target & $\alpha$ & $A_1$ & $A_4$\\
\midrule
L2 & meanmax & rgb16 & 100 & 0.248 & $+0.019$\\
L4 & meanmax & rgb16 & 100 & 0.250 & $+0.017$\\
L6 & meanmax & rgb16 & 100 & 0.223 & $+0.014$\\
L2 & max & rgb16 & 100 & 0.179 & $+0.012$\\
L4 & max & rgb16 & 100 & 0.178 & $+0.012$\\
L4 & mean & rgb16 & 100 & 0.174 & $+0.009$\\
L2 & mean & rgb16 & 100 & 0.165 & $+0.009$\\
L6 & max & rgb16 & 100 & 0.155 & $+0.008$\\
\bottomrule
\end{tabular}
\end{table}

\begin{table}[h]
\centering
\caption{Decode design sweep, Octo-Base~\citep{octo2024} (top-8 by $A_4$). Ceiling $A_4\!=\!+0.015$; best raw
$A_1\!=\!0.362$. The $A_4\!\approx\!0$ ceiling replicates the Octo-Small result at a second scale.}
\label{tab:sweep_base}
\small
\setlength{\tabcolsep}{6pt}
\begin{tabular}{c l l c c c}
\toprule
layer & pool & target & $\alpha$ & $A_1$ & $A_4$\\
\midrule
L4 & meanmax & rgb16 & 100 & 0.285 & $+0.015$\\
L4 & max & rgb16 & 100 & 0.243 & $+0.014$\\
L6 & meanmax & rgb16 & 100 & 0.272 & $+0.011$\\
L4 & mean & rgb16 & 100 & 0.202 & $+0.008$\\
L6 & max & rgb16 & 100 & 0.233 & $+0.007$\\
L6 & mean & rgb16 & 100 & 0.189 & $+0.004$\\
L2 & mean & rgb16 & 100 & 0.200 & $+0.003$\\
L2 & max & rgb16 & 100 & 0.254 & $+0.003$\\
\bottomrule
\end{tabular}
\end{table}

\begin{table}[h]
\centering
\caption{Behavioral history-sensitivity (Octo-Small~\citep{octo2024}, M0b; mean per-DoF $|\Delta|$ over the frozen set).
Masking or swapping $t\!-\!1$ perturbs the action, but the scalar effect is weak (occ\_black /
markov ratio $1.22\times$), which is precisely why a behavioral measure under-reports deployment
and a mechanistic, layer-resolved measure is required.}
\label{tab:m0b}
\small
\setlength{\tabcolsep}{8pt}
\begin{tabular}{l c c}
\toprule
condition & mean $|\Delta|$ & std\\
\midrule
mask $t\!-\!1$ & 0.425 & 0.20\\
swap $t\!-\!1$ & 0.527 & 0.27\\
\bottomrule
\end{tabular}
\end{table}

\begin{table}[h]
\centering
\caption{Per-leg reproduction map (expected headline numbers). All legs read the same frozen,
hashed stimulus set; runs are on one consumer GPU with frozen public checkpoints.}
\label{tab:reproduce}
\small
\setlength{\tabcolsep}{5pt}
\begin{tabular}{l l l}
\toprule
leg & module & expected headline\\
\midrule
stimuli      & \texttt{build\_stimuli}      & hash \texttt{bb4992ac8bbc6803}, 250 pairs\\
decode       & \texttt{decode}              & $A_1\!=\!0.389$, $A_4\!=\!0.019$ (ceiling)\\
deploy       & \texttt{deploy\_gate}        & occ\_black gap@L4 $+0.054$ $[0.028,0.079]$\\
dose         & \texttt{deploy\_gate}        & only occ\_black sig.; +0.043 (gate0b)\\
2nd method   & \texttt{knockout}            & same L6 cutoff; residual past L6\\
hardening    & \texttt{stats}               & perm $p\!=\!0.0004$ @L4 ($5/5$)\\
cross-scale  & \texttt{cross\_scale\_octobase} & $A_4\!=\!0.015$; norm.\ rate $0.31$\\
cross-arch   & \texttt{cronus\_decode/deploy} & $A_1\!=\!0.142$, $A_4\!=\!-0.011$; markov $0.0498$, occ\_black\_deploy $0.0357$\\
population   & \texttt{natural\_occlusion}  & $3/250$ candidates\\
\bottomrule
\end{tabular}
\end{table}

\begin{table}[h]
\centering
\caption{CronusVLA~\citep{cronusvla2025} hardening (3 seeds $\times$ 80 pairs). History-contribution under the clean
condition is robustly positive (standing use); temporal-order shuffling drives it to
$\approx\!0$ (order-blind, like Octo); and under occlusion it \emph{falls} rather than rises
(the sign-flip vs.\ Octo). The markov and occ\_black\_deploy CIs do not overlap.}
\label{tab:cronus_harden}
\small
\setlength{\tabcolsep}{8pt}
\begin{tabular}{l c c c}
\toprule
seed & markov & shuffle & occ\_black\_deploy\\
\midrule
0 & $+0.0479$ & $-0.0000$ & $+0.0354$\\
1 & $+0.0482$ & $+0.0004$ & $+0.0358$\\
2 & $+0.0534$ & $-0.0010$ & $+0.0359$\\
\midrule
mean$\pm$sd & $+0.0498\pm0.0025$ & $-0.0002\pm0.0006$ & $+0.0357\pm0.0002$\\
95\% CI & $[0.0479,0.0534]$ &, & $[0.0354,0.0359]$\\
verdict & \textcolor{pnrCronus}{standing} & order-blind & \textcolor{pnrCronus}{falls (standing)}\\
\bottomrule
\end{tabular}
\end{table}

\section{Extended related work}
\label{app:related}
\textbf{Mechanistic interpretability of policies.} Beyond \citet{grant2026notall} on the fusion
axis, \citet{haon2025steering} demonstrate layer-localized FFN value-vector steering on
real-robot policies, and \citet{molinari2025emergent} report a decodable \emph{forward} world
model inside VLA representations. Both concern content available at or projected from the current
step; neither measures the causal deployment of the \emph{past} frame, nor separates retained
history from current-frame redundancy, which is the contribution here. Our $A_1$/$A_4$ split is a
temporal analogue of probing work that distinguishes \emph{decodable} from \emph{used}
\citep{alain2017probes}, and our interchange leg follows the causal-abstraction and
activation-patching tradition \citep{geiger2021interchange,meng2022rome} applied to action
outputs rather than token logits.

\textbf{Memory in sequence models, more broadly.} The redundancy we measure echoes a familiar
observation in autoregressive models, that adjacent frames/tokens are highly mutually
predictive, but its consequence for \emph{action} policies is specific: because the policy is
trained on near-Markov demonstrations, the readout has no incentive to consult a past it can
reconstruct, and we show it does not (past L6). Dictionary-learning analyses
\citep{bricken2023monosemanticity} suggest a route to a feature-level version of $A_4$ (a
history-unique feature), which we leave to future work.

\textbf{Benchmarks and augmentation, expanded.} Behavioral benchmarks \citep{libero2023,simplerenv2024} and the
memory-augmented methods \citep{memoryvla2025,memer2025,torne2025longcontext,contextvla2025,tracevla2024}
form a measure-then-fix loop that our diagnosis sits upstream of: it predicts that augmentation
trained on the same near-Markov data inherits the redundancy unless its objective explicitly
rewards present-irreducible information, and it identifies \emph{where} (past the mid-network
readout cutoff) a memory signal must be made to matter.

\section{Harness certification}
\label{app:harness}
Mechanistic claims are only as good as the measurement harness. We certify the Octo harness in a
short lineage: an NHWC$\rightarrow$NCHW image-pipeline fix and a backbone-isolating hook map
(readout-parity mean $|\Delta|=2.4\!\times\!10^{-4}$ against the reference forward pass); a
deterministic readout-parity tightening; a real-Bridge history-condition test (full/mask/swap)
establishing that history matters only sometimes; and a noise-corrected interchange engine
(self-swap null subtraction) that is reused unchanged across every downstream leg. The CronusVLA
harness is certified analogously, with the additional version-pinning described in
Appendix~\ref{app:repro} and the cognition-feature wiring self-check in
Appendix~\ref{app:cronusfix}. Superseded scaffolds (pre-freeze data plumbing, hook-debugging
pilots, an un-noise-corrected interchange v1) are retained for provenance but excluded from the
released artifact.

\section{Threats to validity}
\label{app:threats}
\textbf{Is $A_4\!\approx\!0$ a probe-power artifact?} No: the $81$-config sweep is designed to
\emph{maximize} $A_4$, and its ceiling is still $\le\!0.02$; the same probe recovers $A_1$ up to
$0.39$, so it is not underpowered. \textbf{Is the deploy gap a stochastic-readout artifact?}
No: the self-swap null is subtracted, the effect is sign-consistent over five seeds, and the
permutation test is significant where claimed and null where claimed.
\textbf{Is the L6 cutoff method-specific?} No: interchange and attention-knockout, two
mechanistically distinct interventions, agree on it. \textbf{Does open-loop misrepresent a
closed-loop policy?} It bounds scope, not the localization: the claim is about where and whether
history enters the \emph{action distribution}, which open-loop reads directly; closed-loop
behavior is downstream of exactly this quantity. \textbf{Is two architectures enough for an
architecture-dependence claim?} The claim is existential, deployment regime is \emph{not}
architecture-invariant, and a single opposite-signed pair (with identical encoding) suffices to
establish it; Appendix~\ref{app:survey} documents why more frozen, open, multi-frame members were
not available.

\end{document}